\documentclass[journal]{IEEEtran}
\usepackage{times}
\usepackage{epsfig}
\usepackage{amsmath}
\usepackage{amssymb}
\usepackage{graphicx}
\usepackage{cite}
\usepackage[colorlinks,linkcolor=blue]{hyperref} %
\usepackage{color}

\ifCLASSINFOpdf
\else
\fi
\begin{document}
%
\title{The Devil is in the Channels: Mutual-Channel Loss for Fine-Grained Image Classification}

%
%
%

\author{Dongliang~Chang,~Yifeng~Ding,~Jiyang~Xie,~Ayan~Kumar~Bhunia,
Xiaoxu~Li,~Zhanyu~Ma,~Ming~Wu,~Jun~Guo, and~Yi-Zhe~Song 


\thanks{D. Chang, Y. Ding, J. Xie, Z. Ma, M. Wu, J. Guo are with the Pattern Recognition and Intelligent
System Laboratory, School of Information and Communication Engineering, Beijing University of Posts and Telecommunications, Beijing 100876, China.}

\thanks{X. Li is with School of Computer and Communication, Lanzhou University of Technology, Lanzhou 730050, China.}

\thanks{AK Bhunia and Y.-Z. Song  are with the  Centre for Vision, Speech and Signal Processing, University of Surrey, London, United Kingdom.}}

\maketitle

\begin{abstract}
The key to solving fine-grained image categorization is finding discriminate and local regions that correspond to subtle visual traits. Great strides have been made, with complex networks designed specifically to learn part-level discriminate feature representations. In this paper, we show that it is possible to cultivate subtle details without the need for overly complicated network designs or training mechanisms -- a single loss is all it takes. The main trick lies with how we delve into individual feature channels early on, as opposed to the convention of starting from a consolidated feature map. The proposed loss function, termed as mutual-channel loss (MC-Loss), consists of two channel-specific components: a discriminality component and a diversity component. The discriminality component forces all feature channels belonging to the same class to be discriminative, through a novel channel-wise attention mechanism. 
The diversity component additionally constraints channels so that they become mutually exclusive across the spatial dimension. 
The end result is therefore a set of feature channels, each of which reflects different locally discriminative regions for a specific class. The MC-Loss can be trained end-to-end, without the need for any bounding-box/part annotations, and yields highly discriminative regions during inference. Experimental results show our MC-Loss when implemented on top of common base networks can achieve state-of-the-art performance on all four fine-grained categorization datasets (CUB-Birds, FGVC-Aircraft, Flowers-102, and Stanford Cars). Ablative studies further demonstrate the superiority of the MC-Loss when compared with other recently proposed general-purpose losses for visual classification, on two different base networks. Code available at~\url{https://github.com/dongliangchang/Mutual-Channel-Loss}

\end{abstract}


%
\IEEEpeerreviewmaketitle

\section{Introduction}\label{Intro}
\begin{figure}[!h]
  \centering
    \includegraphics[width=1\linewidth]{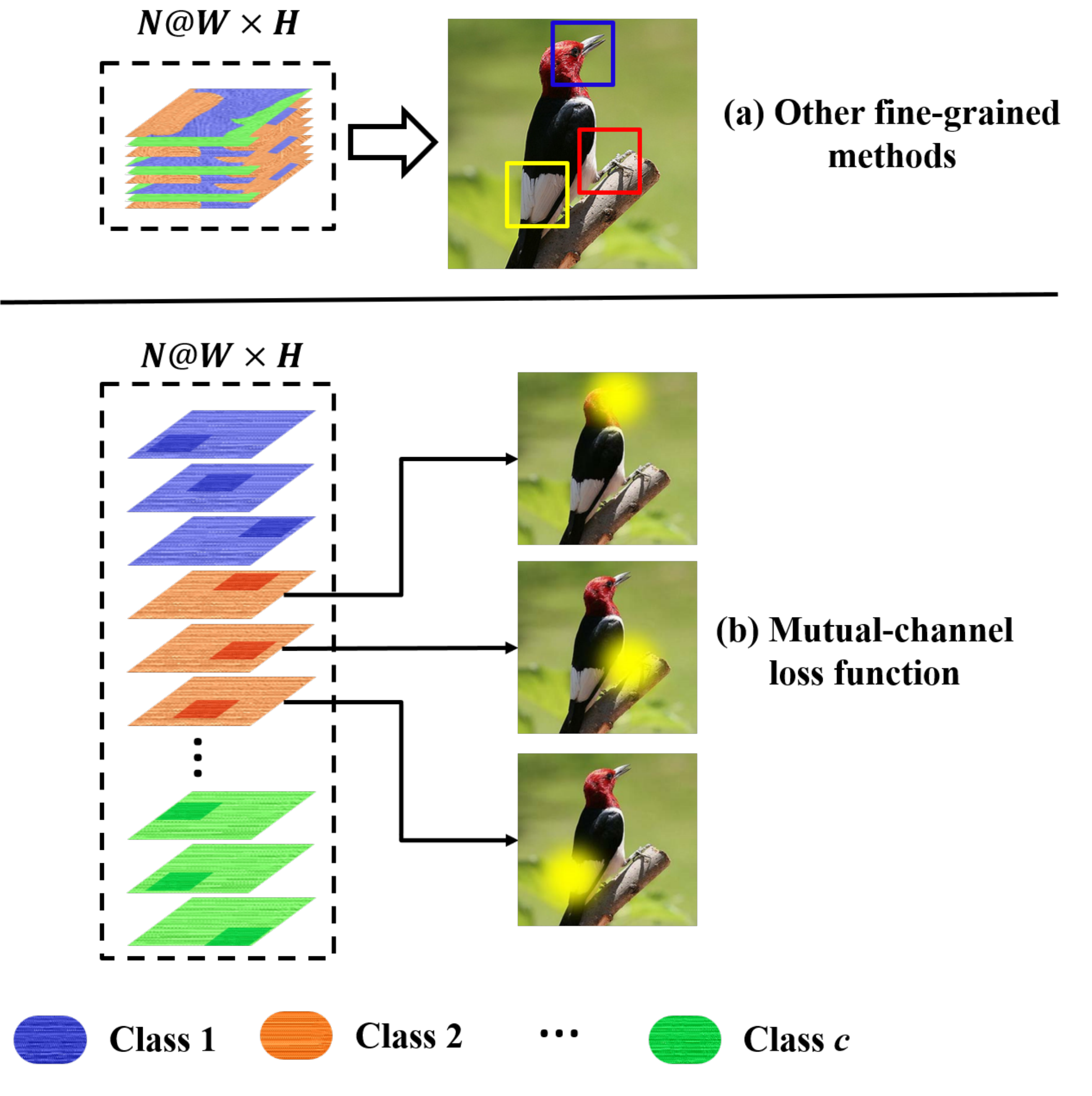}
  \caption{Mutual-channel loss (MC-Loss) where we learn part localized discriminate features directly on channels, without explicit part detection vs. conventional fine-grained classification methods that work on feature maps and with explicit network designs for part detection. We can observe those feature channels after the MC-Loss become class-aligned and each focus on different discriminative regions that roughly correspond to object parts.  
  }
  \label{fig:feature}
\end{figure}

Fine-grained image classification refers to the problem of differentiating sub-categories of a common visual category (\emph{e.g.}, bird species, car models)~\cite{lin2015bilinear}. The task is much harder when compared to conventional category-level classification~\cite{ma2018variational,ma2019insights}, since visual differences between subordinate classes are often subtle and deeply embedded within local discriminative parts. As a result, it has become common knowledge that developing effective methods to extract information from the localized regions that capture subtle differences is the key for solving fine-grained image classification ~\cite{wang2018learning,yang2018learning,zhang2019fine}.

Early works largely relied on manual part annotations, and followed a supervised learning paradigm~\cite{berg2013poof,li2019dual,xie2013hierarchical,branson2014bird,ma2019fine,lei2016fast}. Albeit with decent results reported, it had quickly become apparent that such supervised approaches are not scalable. This is because expert human annotations can be cumbersome to obtain and are often error-prone~\cite{volkmer2005web}. More recent research has therefore concentrated on realizing parts in an unsupervised fashion ~\cite{lin2015bilinear,song2017parameter,zhang2016picking,Peng2018Object,cong2017iterative,wei2019adversarial,deng2019unsupervised}. These approaches have been shown to yield performances on par or even exceeding those that relied on manual annotations, owing to their ability of mining discriminative parts that are otherwise missing or inaccurate in human labelled data. Again, the main focus is placed on how best to locate discriminative object parts. Increasingly more complicated networks have been proposed to perform part learning, mainly to compensate for the lack of annotation data. Two main components can be typically identified amongst these approaches: (i) a network component to explicitly perform part detection, and (ii) a way to ensure that features learned are maximally discriminative. Most recent work on fine-grained classification~\cite{wang2018learning,yang2018learning,zheng2017learning,fu2017look,lei2019semi} has shown state-of-the-art performance by simultaneously exploring these two components, cultivating their complementary properties.

In this paper, we follow the same motivation as above~\cite{zhang2016picking,wang2015multiple,zheng2017learning} to address the unique challenges of fine-grained classification. We importantly differ in 
not attempting to introduce any explicit network components for discriminate part discovery. Instead, we ask if it is even possible to simultaneously achieve both discriminative features for learning and part localization, with just a single loss. This design choice has a few salient advantages over prior art: (i) it does not introduce any extra network parameters, making the network easier to train, and (ii) it can in principle be applied to any existing or future network architecture. The key insight lies with how we delve into feature channels early on, as opposed to learning fine-grained part-level features on feature maps directly.

More specifically, we assume a fixed number of feature channels to represent each class. It follows that instead of applying constraints on the final feature maps, we impose a loss directly on the channels, so that all the feature channels belonging to the same class are (i) discriminative, \emph{i.e.}, they each contribute to discriminating the class from others, and (ii) mutually exclusive, \emph{i.e.}, each channel can attend to different local regions/parts. The end result is therefore, a set of feature channels that are class-aligned, each being discriminative on mutually distinct local parts. Figure~\ref{fig:feature} offers a visualization. To the best of our knowledge, this is the first time that a single loss is proposed for fine-grained classification that does not require any specific network designs for partial localization.

Our loss is termed mutual-channel loss, MC-Loss in short. It has two components that work synergistically for fine-grained feature learning. Firstly, a discriminality component is introduced to enforce all feature channels corresponding to a class to be discriminative on their own, before being fused. A novel channel attention mechanism is introduced, whereby during training a fixed percentage of channels is randomly masked out, forcing the remaining channels to become discriminative for a given class. We then apply  cross-channel max pooling~\cite{goodfellow2013maxout} to fuse the feature channels and produce the final feature map which is now class-aligned and optimally discriminative.

Although every feature channel is now discriminative against a class, there is still no guarantee that most discriminative parts will be localized. This leads us to introduce the second component of our loss function, the diversity component. This component is specifically designed so that each channel within a group will attend to mutually distinct local parts. We achieve this goal by asking for maximum spatial de-correlation across channels belonging to the same class. This can be conveniently implemented by (again) applying cross-channel max pooling, then asking for maximum spatial summation. Ultimately, this is done to ensure as many discriminative parts are attended to as possible, therefore helping with fine-grained feature learning. Note that the diversity component would not work without its discriminative counterpart, since otherwise not all channels would be discriminative making localization much harder.

Extensive experiments are carried out on four commonly used fine-grained categorization datasets, CUB-$200$-$2011$~\cite{wah2011caltech}, FGVC-Aircraft~\cite{maji2013fine}, Flowers-$102$~\cite{nilsback2008automated}, and Stanford Cars~\cite{krause20133d}. The results show that our method can outperform the current the state-of-the-arts by a significant margin. Ablative studies are further conducted to draw insights towards each of the proposed loss components, and hyper-parameters.

\begin{figure*}[!t]
  \centering
    \includegraphics[width=0.9\linewidth]{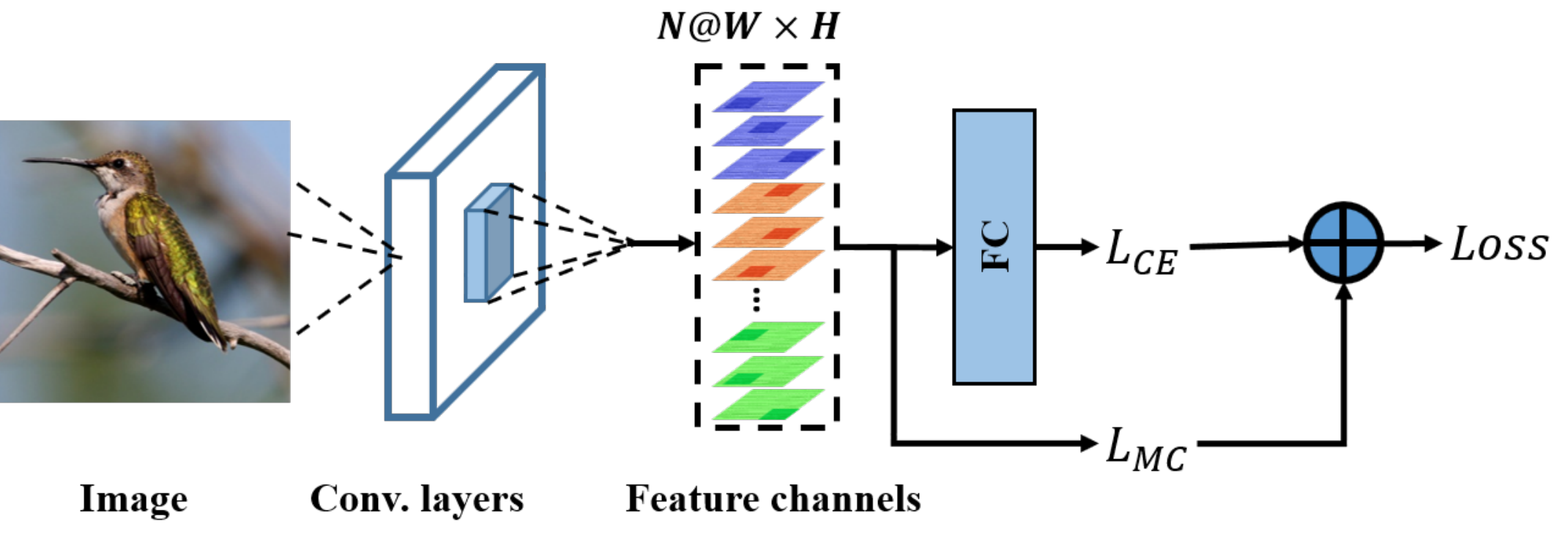}
  \caption{The framework of a typical fine-grained classification network where MC-Loss is used. The MC-Loss function considers the output feature channels of the last convolutional layer as the input and gathers together with the cross-entropy (CE) loss function using a hyper-parameter $\mu$.}
  \label{fig:network}
\end{figure*}
\section{Related Work}\label{RelatedWork}

In this section, we briefly review previous works regarding both fine-grained image classification and relevant loss functions for similar purposes.
\subsection{Fined-Grained Image Classification}

Some of the earlier works~\cite{berg2013poof,chai2013symbiotic,xie2013hierarchical,min2017new} take advantages of bounding-box/part annotations, as an additional information for both training and testing. However, expert annotations are hard to source and can be prone to human error, and thus it hinders practical deployment in the wild scenarios. To address this issue, some other works~\cite{branson2014bird,zhang2014part} use annotations only during training and utilize a part-detection module during testing.

Recently, some frameworks employ a more general architecture that can localize discriminative parts within an image without any extra supervision from part annotations, and thus it makes the fine-grained image classification more feasible in real-world scenarios. Wang~\emph{et al.}~\cite{wang2018learning} claimed that improving mid-level convolutional feature representation can bring significant advantages for part-based fine-grained classification. This is accomplished by introducing a bank of discriminative filters in the classical convolutional neural networks (CNNs) architecture and it can be trained in an end-to-end fashion. Authors in~\cite{dubey2018pairwise} presented a new procedure, called pairwise confusion (PC), in order to improve the generalization for fine-grained image classification task by encouraging confusion in the output activations and forcing the model to focus on local discriminative features of the objects rather than the backgrounds. Meanwhile, Yang~\emph{et al.}~\cite{yang2018learning} proposed a novel multi-agent cooperative learning scheme which learns to identify the discriminative regions in the image in a self-supervised way.

Despite all these improvements, part-based methods have difficulties in modelling the specific features of an image because of the complicated relationship that exists between the different distinct parts. In order to handle this complex interaction, some approaches encode higher-order statistics of convolutional features and extract compact holistic representations.  Lin~\emph{et al.}~\cite{lin2015bilinear} added a bilinear pooling behind the dual CNNs to obtain discriminative feature representation of the whole image. 
As an extension of bilinear pooling, Cui~\emph{et al.}~\cite{cui2017kernel} proposed a deep kernel pooling method that captures the high-order, non-linear feature interactions via compact and explicit feature mapping. A higher-order integration of hierarchical convolutional features has been introduced into an end-to-end framework to derive rich representation of the local parts at different scales for fine-grained image classification~\cite{cai2017higher}.
The very recent work by Zheng~\emph{et al.}~\cite{zheng2017learning} is perhaps the closest to our work since they also operated at channel-level. They designed a multi-attention convolutional neural network (MA-CNN) to jointly learn discriminative parts and fine-grained feature representation on each channel, which then got aggregated later on to construct the final fine-grained features. 

Without exception, all previous approaches incur network changes to achieve part localization and/or discriminative feature learning. This is distinctively different to our approach of achieving the same via a single loss function. 


\subsection{Loss Functions in CNNs}

A recent trend has been noticed in the computer vision community towards designing task-specific loss functions to reinforce the CNNs with strong discriminative information. Intuitively, the extracted features are most discriminative when their intra-class compactness and inter-class separability are simultaneously maximized, \emph{i.e.} the Fisher criterion. Wen~\emph{et al.}~\cite{wen2016discriminative} proposed the center loss to obtain the highly discriminative features for robust recognition by minimizing the inter-class distance of deep features. Liu~\emph{et al.}~\cite{liu2017sphereface} introduced the A-softmax loss to learn angularly discriminative features for image classification on a deep hypersphere embedding manifold. Wang~\emph{et al.}~\cite{wang2018cosface} embraced the idea of the Fisher criterion and proposed the large margin cosine loss (LMCL) to learn highly discriminative deep features for image recognition.

In addition, there are some works that focus on the effective use of training data. Lin~\emph{et al.}~\cite{lin2017focal} proposed the focal loss, a modified cross-entropy (CE) loss, in order to emphasize learning on hard samples and down-weight well-classified samples. Wan~\emph{et al.}~\cite{wan2018rethinking} proposed the large-margin Gaussian mixture (L-GM) loss by assuming a Gaussian mixture distribution for the deep features on the training set, which boosts a more effective discrimination of out-of-domain inputs.

Although all the aforementioned loss functions can obtain discriminative features to an extent, they do not explicitly encourage the network to focus on the localized discriminative regions. In contrast, our proposed MC-Loss function enforces the network to discover multiple discriminative regions, which also alleviates the need of complicated network designs unlike\cite{zhang2016picking,wang2015multiple,zheng2017learning}, and thus it makes our framework easy-to-implement and easy-to-interpret.

\begin{figure*}[!t]
  \begin{center}
    \includegraphics[width=1.0\linewidth]{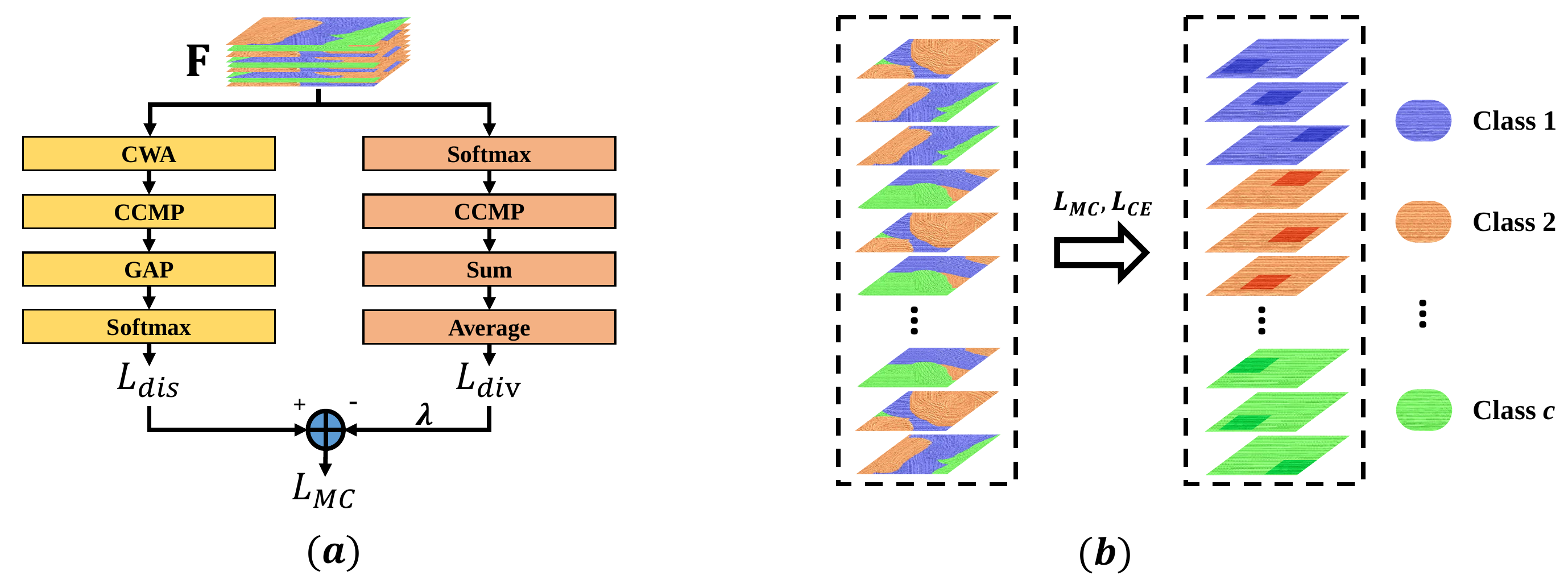}
  \end{center}
  \caption{(a) Overview of the MC-Loss. The MC-Loss consists of (i) a discriminality component (left) that makes $\mathcal{F}$ to be class-aligned and discriminative, and (ii) a diversity component (right) that supervises the feature channels to focus on different local regions. (b) Comparison of feature maps before (left) and after (right) applying MC-Loss, where feature channels become class aligned, and each attending to different discriminate parts.~\textbf{Please refer to Section~\ref{Methods} for details.}
  }
  \label{fig:MCloss}
\end{figure*}

\section{The Mutual-Channel Loss (MC-Loss)}\label{Methods}

In this section, we propose the mutual-channel loss (MC-Loss) function to effectively navigate the model focusing on different discriminative regions without any fine-grained bounding-box/part annotations.

The network combined with the proposed MC-Loss in the training step is shown in Figure~\ref{fig:network}. Given an input image, it first extracts the feature maps by feeding the image into a base network;~\emph{e.g.}, VGG$16$~\cite{simonyan2014very} or ResNet$18$~\cite{he2016deep}. Let the extracted feature maps be denoted as $\mathcal{F} \in R^{N\times W\times H}$, with height $H$, width $W$,  and number of channels $N$. In the proposed MC-Loss, we need to set the value of $N$ equals to $c\times\xi$, where $c$ and $\xi$ indicate the number of classes in a dataset and the number of feature channels used to represent each class, respectively. Note that $\xi$ is a scalar hyper-parameter and empirically larger than 2. The  $n^{th}$ vectored feature channel of $\mathcal{F}$ is represented as $\mathcal{F}_n\in R^{W\!H}, n=1,2,\cdots,N$. Please note that we reshape each channel matrix of $\mathcal{F}$ of dimension $W\times H$ to a vector of size $W$ \!times $H$, \emph{i.e.} $W\!H$. The grouped feature channels corresponding to $i^{th}$ class is indicated by $\mathbf{F}_i\in R^{\xi\times W\!H}, i=0,1,\cdots,c-1$. Mathematically, it can be represented as

\begin{equation}
  \mathbf{F}_i=\left\{\mathcal{F}_{i\times\xi +1}, \mathcal{F}_{i\times\xi +2}, \cdots, \mathcal{F}_{i\times\xi+\xi}\right\}.
\end{equation}

Subsequently, $\mathbf{F}=\left\{\mathbf{F}_0,\mathbf{F}_1,\cdots,\mathbf{F}_{c-1}\right\}$ enters into two streams of the network with two different sub-losses tailored for two distinct goals. In Figure~\ref{fig:network}, the cross-entropy stream considers $\mathbf{F}$ as the input to a fully connected (FC) layers with traditional CE loss $L_{CE}$. Here, the cross-entropy loss encourages the network to extract informative features which mainly focus on the~\emph{global discriminative regions}. On the other side, the MC-Loss stream supervises the network to spotlight \emph{different local discriminative regions}. The MC-Loss is then added to the CE loss with the weight of $\mu$ in the training step. Thus, the total loss function of the whole network can be defined as

\begin{equation}
  Loss(\mathbf{F})=L_{CE}(\mathbf{F})+\mu\times L_{MC}(\mathbf{F}).
 \label{total_loss}
\end{equation}

Furthermore, the MC-Loss is a weighted summation of one discriminality  component ${L}_{dis}$ and  another  diversity component ${L}_{div}$. We define the MC-Loss as

\begin{equation}
  L_{MC}(\mathbf{F})={L}_{dis}(\mathbf{F})-\lambda\times{L}_{div}(\mathbf{F}).
    \label{mcloss}
\end{equation}


\subsection{The Discriminality Component}

In our framework, each class is represented by a certain number of grouped feature channels.
The discriminality component enforces the feature channels to be class-aligned and each feature channel corresponding to a particular class should be discriminative enough. The discriminality component $L_{dis}$ can be represented as

\begin{small}
  \begin{equation}\label{eq:L_dis}
    {L}_{dis}(\mathbf{F})={L}_{CE}\big(\boldsymbol{y},\underbrace{\frac{{\left[e^{g(\mathbf{F}_0)}, e^{g(\mathbf{F}_1)}, \cdots, e^{g(\mathbf{F}_{c-1})}\right]^{\text{T}}}} {\sum_{i=0}^{c-1}e^{g(\mathbf{F}_i)}}}_{\text{Softmax}}\big),
  \end{equation}
\end{small}

\noindent where $g(\cdot)$ is defined as

\begin{small}
  \begin{equation}
    g(\mathbf{F}_i)=\underbrace{\frac{1}{W\!H}\sum_{k=1}^{W\!H}}_{\text{GAP}} \underbrace{\max_{j=1,2,\cdots,\xi}\vphantom{\sum_{n=1}^{N}}}_{\text{CCMP}} \underbrace{\left[M_i\cdot\mathbf{F}_{i,j,k}\right]\vphantom{\sum_{n=1}^{N}}}_{\text{CWA}},
  \end{equation}
\end{small}

\noindent where the GAP, the CCMP, and the CWA are short notations for global average pooling, cross-channel max pooling, and channel-wise attention,  respectively. ${L}_{CE}(\cdot, \cdot)$ is the cross-entropy loss between the ground-truth class label $\boldsymbol{y}$ and the output of GAP. $M_i=diag(\textbf{Mask}_i)$, where $\textbf{Mask}_i\in R^{\xi}$ is a $0$-$1$ mask with randomly $\left\lfloor\frac{\xi}{2}\right\rfloor$ zero(s). The  $\left\lceil\frac{\xi}{2}\right\rceil$ ones and operation $diag(\cdot)$ put a vector on the principle diagonal of a diagonal matrix. The left block in Figure~\ref{fig:MCloss}(a) shows the flow diagram of the discriminality component.

\noindent \textbf{CWA:} While in case of traditional CNNs, trained with the classical CE loss objective, a certain subset of feature channels contain discriminative information, we here propose channel-wise attention operation to enforce the network to equally capture discriminative information in all $\xi$ channels corresponding to a particular class.  Unlike other channel-wise-attention design~\cite{chen2017sca} that intends to assign higher priority to the discriminative channels using soft-attention values, we assign random binary weights to the channels and stochastically select a few feature channels from every feature group $\mathbf{F}_i$ during each iteration, thus explicitly encouraging every feature channel to contain sufficient discriminative information. This process could be visualized as a random channel-dropping operation. Please note that the CWA is used only during training and that the whole MC-Loss branch is not present at the time of inference. Therefore, the classification layer receives the same input feature distributions during both training and inference.

\noindent \textbf{CCMP:} Cross-channel max pooling~\cite{goodfellow2013maxout} is used to compute the maximum response of each element across each feature channel in $\mathbf{F}_i$ corresponding to a particular class, and thus it results into a one dimensional vector of size $W\!H$ concurring to a particular class. Note that the cross-channel average pooling (CCAP) is an alternative of the CCMP, which only substitutes the max pooling operation by the average pooling. However, the CCAP tends to average each element across the group which may suppress the peaks of feature channels, \emph{i.e.}, attentions of local regions. On the contrary, the CCMP can preserve these attentions, and is found to be beneficial for fine-grained classification.

\noindent \textbf{GAP:} Global average pooling~\cite{lin2013network} is used to compute the average response of each feature channel, resulting in a $c$-dimensional vector where each element corresponds to one individual class.

Finally, we use the CE loss function $L_{CE}$ to compute the dissimilarity between the ground-truth labels and the predicted probabilities given by the softmax function behind the GAP operation.

\subsection{The Diversity Component}

The diversity component is an approximated distance measurement for feature channels to calculate the total similarity of all the channels. It is cheaper in computation with a constant complexity than other commonly used measurements like Euclidean distance and Kullback-Leibler divergence with a quadratic complexity. The diversity component illustrated along the right block of Figure \ref{fig:MCloss}(a) drives the feature channels in a group $\mathbf{F}_i$ to become different from each other via training. In other words, different feature channels of a class should focus on different regions of the image, rather than all the channels focusing on the most discriminative region. Thus, it reduces the redundant information by diversifying the feature channels from  every group and helps to discover different discriminative regions with respect to every class in an image. This operation can be interpreted as a cross-channel de-correlation in order to capture details from different salient regions of an image. After the softmax, we impose supervision directly at the convolutional filters by introducing a CCMP followed by a spatial-dimension summation to measure the degree of intersection. The diversity specific loss component $L_{div}$ can be defined as

\begin{equation}
  {L}_{div}(\mathbf{F})=\frac{1}{c}\sum_{i=0}^{c-1}h(\mathbf{F}_i),
  \label{equ:ldiv}
\end{equation}

\noindent where $h(\cdot)$ is defined as 

\begin{equation}
  h(\mathbf{F}_i)=\sum_{k=1}^{W\!H} \underbrace{\max_{j=1,2,\cdots,\xi}\vphantom{\sum_{n=1}^{N}}}_{\text{CCMP}} \underbrace{\left[\frac{e^{\mathbf{F}_{i,j,k}}}{\sum_{k'=1}^{W\!H}e^{\mathbf{F}_{i,j,k'}}}\right]\vphantom{\sum_{n=1}^{N}}}_{\text{Softmax}}.
  \label{diver_loss}
\end{equation}

\noindent The function softmax is a normalization on spatial dimensions and the CCMP here plays the same role as it does in the discriminality component.

\begin{figure}[!t]
  \begin{center}
    \includegraphics[width=0.8\linewidth]{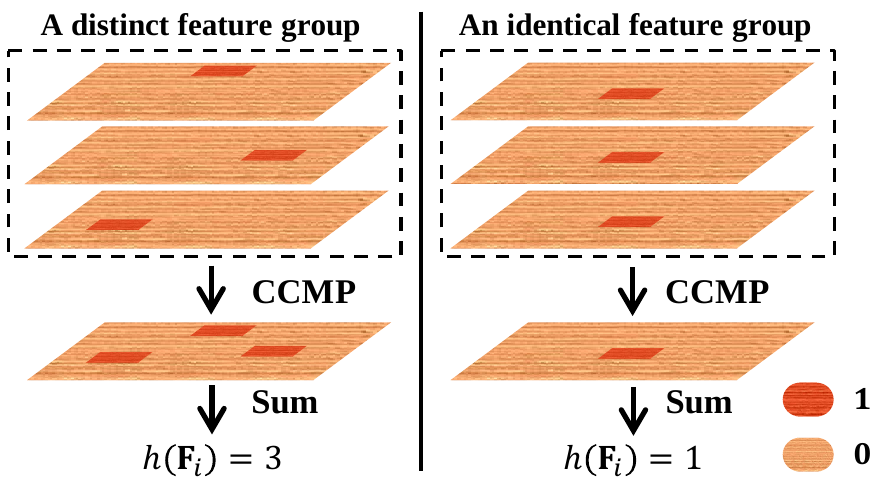}
  \end{center}
  \caption{A graphical explanation of the diversity component.}
  \label{fig:explaindivloss}
\end{figure}

The upper-bound of $L_{div}$ is equal to $\xi$ in the case of $\xi$ extremely different feature maps which means that they focus on different local regions, while the lower-bound is $1$ facing $\xi$ same feature maps in $\mathbf{F}_i$ which need to be optimized clearly shown in Figure~\ref{fig:explaindivloss}. In Figure~\ref{fig:explaindivloss}, assuming that each feature channel is one-hot normalized by softmax,  $h(\cdot)$ would response to the upper-bound $3$ ($\xi=3$) if each feature channel has the one in distinct locations, \emph{i.e.},  focusing on different local regions. Conversely, if obtaining identical feature channels, $h(\cdot)$ would response to the lower-bound $1$. Ideally, we intend to maximize the $L_{div}$ term and thus it justifies the minus sign in Equation \ref{mcloss}. A point is to be noted that the diversity component cannot work alone for classification, it acts as a regularizer on the top of discriminality loss to implicitly discover different discriminative regions in an image. Intuitively, we will discuss the availability of the diversity component in visualization results in Section~\ref{Ablation}.



\begin{table}[!t]
  \centering
  \small
  \caption{Statistics of datasets.}
    \begin{tabular}{|c|c|c|c|}
    \hline
    Datasets         & \#Category & \#Training & \#Testing \\
    \hline
    \hline
    CUB-$200$-$2011$ & $200$      &$5994$      &$5794$\\
    FGVC-Aircraft    & $100$      &$6667$      &$3333$\\
    Stanford Cars    & $196$      &$8144$      &$8041$\\
    Flowers-$102$    & $102$      &$2040$      &$6149$\\
    \hline
    \end{tabular}%
  \label{tab:datasets}
\end{table}

\begin{table}[!t]
  \centering
  \small
  \caption{$\xi$ value assignment while \textbf{using the pre-trained VGG$16$/ResNet$50$} with $512$/$2048$ feature channels.}
    \begin{tabular}{|c|c|c|}
    \hline
    Datasets              &  $2$/$10$ feature channels   &  $3$/$11$ feature channels \\
    \hline
    \hline
    CUB-$200$-$2011$      &$88$/$152$                    &$112$/$48$ \\
    Stanford Cars         &$76$/$108$                    &$120$/$88$  \\
    \hline

    \multicolumn{1}{r}{}  & \multicolumn{1}{r}{}         & \multicolumn{1}{r}{} \\
    \hline 
    Datasets              &$5$/$20$  feature channels    & $6$/$21$ feature channels \\
    \hline
    \hline
    FGVC-Aircraft         &$88$/$52$                     &$12$/$48$\\
    Flowers-$102$         &$100$/$94$                    &$2$/$8$ \\
    \hline
    \end{tabular}%
  \label{tab:featuremaps}%
\end{table}%

\begin{table}[!t]
  \centering
  \caption{Experimental results ($\%$) on Flowers-$102$ dataset using the \textbf{pre-trained VGG$16$ and ResNet$50$}.}
    \footnotesize
    \begin{tabular}{|c|c|c|}
    \hline
    Method                                             & Base Model              & Flowers-$102$ \\
    \hline
    \hline
    Det.+seg. (CVPR$13$~\cite{angelova2013efficient})  & SVM                     &$80.7$ \\
    Overfeat (CVPR$14$ workshop~\cite{sharif2014cnn})  & Overfeat                &$86.8$ \\
    B-CNN (ICCV$15$~\cite{lin2015bilinear})            & VGG$16$                 &$92.5$ \\
    Selective joint FT (CVPR17~\cite{ge2017borrowing}) & ResNet$152$             &$95.8$ \\
    PC (ECCV$18$~\cite{dubey2018pairwise} )            & B-CNN                   &$93.7$ \\
    PC (ECCV$18$~\cite{dubey2018pairwise} )            & DenseNet$161$           &$91.2$ \\
    \hline
    \hline
    MC-Loss                                            & VGG$16$                 &$96.1$ \\
    MC-Loss                                            & ResNet$50$              &$96.8$ \\
    MC-Loss                                            & B-CNN                   &$\textbf{97.7}$ \\
    \hline
    \end{tabular}%
  \label{tab:SOTA_flowers102}

\end{table}%

\begin{table*}[!t]
  \centering
  \footnotesize
  \caption{Experimental results ($\%$) on CUB-$200$-$2011$, FGVC-Aircraft, and Stanford Cars datasets, respectively with \textbf{pre-trained VGG$16$ and ResNet$50$}. The best and second best results are respectively marked in \textbf{bold} and \underline{\emph{ITALIC}} fonts.}
    \begin{tabular}{|c|c|c|c|c|c|}
    \hline
    Method                                       & Base Model     & CUB-$200$-$2011$              & FGVC-Aircraft                 & Stanford Cars    & Model Component \\
    \hline
    \hline
    FT VGGNet (CVPR$17$~\cite{wang2018learning}) & VGG$19$                    & $77.8$                        & $84.8$                        & $84.9$            & C+A \\
    FT ResNet (CVPR$18$~\cite{wang2018learning}) & ResNet$50$                 & $84.1$                        & $88.5$                        & $91.7$            & D+A \\
    B-CNN (ICCV$15$~\cite{lin2015bilinear})      & VGG$16$                    & $84.1$                        & $84.1$                        & $91.3$            & $2$B +A \\
    KA (ICCV$17$~\cite{cai2017higher})           & VGG$16$                    & $85.3$                        & $88.3$                        & $91.7$            & B+A+Conv.($1$,$1$) \\
    KP (CVPR$17$~\cite{cui2017kernel})           & VGG$16$                    & $86.2$                        & $86.9$                        & $92.4$            & B+ kernel pooling+A \\
    MA-CNN (ICCV$17$~\cite{zheng2017learning})   & VGG$19$                    & $86.5$                        & $89.9$                        & $92.8$            & C + $3$A + channel grouping layers \\
    PC (ECCV$18$~\cite{dubey2018pairwise})       & B-CNN                      & $85.6$                        & $85.8$                        & $92.5$            & $2$B+A \\
    PC (ECCV$18$~\cite{dubey2018pairwise})       & DenseNet$161$              & $86.9$                        & $89.2$                        & $92.9$            & $2$E + $2$A \\
    DFL-CNN (CVPR$18$~\cite{wang2018learning})   & VGG$16$                    & $86.7$                        & $92.0$                        & $93.8$            & B+$2$A+Conv.($1$,$1$) \\
    DFL-CNN (CVPR$18$~\cite{wang2018learning})   & ResNet$50$                 & $87.4$                        & $91.7$                        & $93.1$            & D+$2$A+Conv.($1$,$1$) \\
    NTS-Net (ECCV$18$~\cite{yang2018learning})   & ResNet$50$                 & $87.5$                        & $91.4$                        & $\underline{\textit{93.9}}$   & D+$3$A+$6$Conv.($3$,$3$) \\
    WPS-CPM (CVPR$19$~\cite{Ge_2019_CVPR})       & GoogLeNet + ResNet$50$     & $\textbf{90.4}$               &      -                        & -                  & GoogleNet + D+A \\
    TASN (CVPR$19$~\cite{Zheng_2019_CVPR})       & ResNet$50$                 & $\underline{\textit{87.9}}$   &      -                        & $93.8$                  &  D+A \\
    \hline
    \hline
    MC-Loss                                      & VGG16                      & $78.7$                        & $91.0$                        & $92.8$ & B+A \\
    MC-Loss                                      & ResNet50                   & $87.3$                        & $\underline{\textit{92.6}}$   & $93.7$ & D+A \\   
    MC-Loss                                      & B-CNN                      & $86.4$                        & $\textbf{92.9}$               & $\textbf{94.4}$ & $2$B+A \\
    \hline
    \end{tabular}%
  \label{tab:SOTA}%
\end{table*}%

\begin{table*}[!t]
  \centering
  \small
  \caption{Influence of feature channel number on four fine-grained image classification datasets (trained from scratch). $\xi$=$i$ means each category has $i$ feature channels.}
    \begin{tabular}{|c|c|c|c|c|c|}
    \hline
    Method              & Base Model   & CUB-$200$-$2011$ & FGVC-Aircraft     & Stanford Cars   & Flowers-$102$ \\
    \hline
    \hline
    MC-Loss ($\xi$=$1$) & VGG$16$      & $58.80$          & $82.08$           & $84.88$          & $69.99$ \\
    MC-Loss ($\xi$=$2$) & VGG$16$      & $62.11$          & $88.66$           & $90.61$          & $81.98$ \\
    MC-Loss ($\xi$=$3$) & VGG$16$      & $\textbf{65.98}$ & $\textbf{89.20}$  & $\textbf{90.85}$ & $\textbf{83.23}$ \\
    MC-Loss ($\xi$=$5$) & VGG$16$      & $64.39$          & $89.01$           & $90.80$          & $82.84$ \\
    MC-Loss ($\xi$=$6$) & VGG$16$      & $63.08$          & $88.22$           & $89.82$          & $81.26$ \\
    \hline
    \hline
    MC-Loss ($512$)     & VGG$16$      &$ 62.27$          & $88.46$           &$ 90.78 $         & $82.57$ \\
    \hline
    \end{tabular}%
  \label{tab:mis_match}%
\end{table*}%

\section{Experimental Results and Discussions}\label{Experiments}

In this section, we evaluate the performance of our proposed method on the fine-grained image classification task. Firstly, the datasets and the implementation details are introduced in Section~\ref{Datasets} and~\ref{Details}, respectively.  Subsequently, the classification accuracy comparisons with other state-of-the-art methods are then provided in Section~\ref{SOTA}. In order to illustrate the advantages of different loss-components and design choices, a comprehensive ablation study is provided in Section~\ref{Ablation}.

\subsection{Datasets}\label{Datasets}

We evaluate the proposed MC-Loss on four widely used fine-grained image classification datasets, namely Caltech-UCSD-Birds (CUB-$200$-$2011$)~\cite{wah2011caltech}, FGVC-Aircraft~\cite{maji2013fine}, Stanford Cars~\cite{krause20133d}, and Flowers-$102$~\cite{nilsback2008automated}. The detailed summary of the datasets are provided in Table~\ref{tab:datasets}. In order to keep consistency with other datasets, where datasets are divided into training and test set only, we consider both training and validation sets for training in case of Flowers-$102$ dataset. Moreover, we only use the category labels in our experiments.

\subsection{Implementation Details}\label{Details}

The foremost important thing to be noted is that the number of channels in the output feature maps extracted from a pre-trained VGG$16$ (ResNet$50$), is fixed at $512$ ($2048$). Say for example, if want to fix $\xi = 3$ uniformly for every class, this would require $600$, $300$, $588$, and $306$ feature channels for CUB-$200$-$2011$ (with 200 classes), FGVC-Aircraft (with 150 classes),  Stanford Cars (with 196 classes), and Flowers-$102$ datasets (with 102 classes), respectively. This is not feasible with the pre-trained VGG$16$ (ResNet$50$) since the number of feature channel is fixed at $512$  ($2048$). On the other side, we intend to explore the pre-trained rich discriminative features of the VGG$16$ (ResNet$50$) that is learned on large ImageNet dataset and we fine-tune the pre-trained models with our proposed loss function in Equation \ref{total_loss}. Therefore, we assign $\xi$ non-uniformly in order to serve the purpose of using pre-trained VGG$16$  (ResNet$50$). Say for example, when we fine-tune the VGG$16$ pre-trained on the ImageNet classification dataset, we assign $2$ feature channels for each of the first $88$ classes and the rest $112$ classes are modelled with $3$ feature channels in case of CUB-$200$-$2011$ dataset. Please refer to Table \ref{tab:featuremaps} for details.

In order to compare our proposed loss function with other state-of-the-art methods (see Table \ref{tab:SOTA_flowers102} and \ref{tab:SOTA}), we resize every image to a size of $448\times448$ (following others), then extract features using the VGG$16$ (ResNet$50$), and the B-CNN \cite{lin2015bilinear} based on a VGG$16$ model pre-trained on the ImageNet classification dataset. We use Stochastic Gradient Descent optimizer and batch normalization as the regularizer. The learning rate of the pre-trained feature extraction layers are kept as $1\times10^{-4}$, while the learning rate of fully connected layers is initially set at $0.01$ and multiplied by $0.1$ at $150^{th}$ and $225^{th}$ epoch, successively. We train our model for $300$ epochs and weight decay value is kept as  $5\times10^{-4}$. Furthermore, we set the hyper-parameters of the MC-Loss as $\mu$ =$0.005$ and $\lambda$=$10$.

\begin{table*}[!t]
  \centering
  \small
  \caption{Comparisons of classification accuracies ($\%$) with different loss functions using the VGG$16$ and the ResNet$18$ as backbone architecture (trained from scratch). The best and the second best results are respectively marked in \textbf{bold} and \underline{\emph{ITALIC}} fonts.}
    \begin{tabular}{|c|c|c|c|c|c|}
    \hline
    Method                                   & Base Model                                 & CUB-$200$-$2011$                        & FGVC-Aircraft                                                & Stanford Cars                             & Flowers-$102$ \\
    \hline
    \hline
    CE Loss                                  & VGG$16$ / ResNet$18$                       & $28.53$ / $45.70$                       & $82.90$ / $79.90$                                            & $76.59$ / $79.12$                         & $40.90$ / $65.75$  \\
    Center Loss~\cite{wen2016discriminative} & VGG$16$ / ResNet$18$                       & $51.38$ / $\underline{\textit{50.26}}$  & $\underline{\textit{88.26}}$ / $\underline{\textit{83.86}}$  & $\underline{\textit{89.27}}$ / $81.84$    & $62.53$ / $69.51$  \\
    A-softmax Loss~\cite{liu2017sphereface}  & VGG$16$ / ResNet$18$                       & $\underline{\textit{60.79}}$ / $49.67$  & $88.15$ / $82.42$                                            & $88.71$ / $\underline{\textit{82.15}}$    & $62.34$ / $50.56 $ \\
    Focal Loss~\cite{lin2017focal}           & VGG$16$ / ResNet$18$                       & $31.12$ / $47.67$                       & $80.85$ / $80.47$                                            & $77.02$ / $79.75$                         & $48.19$ / $66.87$  \\
    COCO Loss~\cite{liu2017rethinking}       & VGG$16$ / ResNet$18$                       & $48.31$ / $46.01$                       & $86.41$ / $80.02$                                            & $67.27$ / $72.38$                         & $63.31$ / $66.76$ \\
    LGM Loss~\cite{wan2018rethinking}        & VGG$16$ / ResNet$18$                       & $28.14$ / $44.91$                       & $87.49$ / $80.98$                                            & $71.27$ / $74.37$                         & $57.78$ / $66.84$  \\
    LMCL Loss~\cite{wang2018cosface}              & VGG$16$ / ResNet$18$                       & $41.11$ / $46.01$                       & $86.17$ / $78.52$                                            & $49.57$ / $71.17$                         & $\underline{\textit{66.43}}$ / $\underline{\textit{67.72}}$ \\
    \hline
    \hline
    MC-Loss                                  & VGG$16$ / ResNet$18$                       & $\textbf{65.98}$ / $\textbf{59.41}$     & $\textbf{89.2}$ / $\textbf{85.57}$                            & $\textbf{90.85}$  / $\textbf{87.47}$     & $\textbf{83.23}$ / $\textbf{79.54}$ \\
    \hline
    
    \end{tabular}%
  \label{tab:table_loss_function}%
\end{table*}%

\begin{table*}[!t]
  \centering
  \caption{Ablation study of the MC-Loss (trained from scratch) on four fine-grained image classification datasets.}
  \small
    \begin{tabular}{|c|c|c|c|c|c|}
    \hline
    Method                              & Base Model    & CUB-$200$-$2011$    & FGVC-Aircraft      & Stanford Cars       & Flowers-$102$ \\
    \hline
    \hline
    MC-Loss                             & VGG$16$       & $\textbf{65.98}$    & $\textbf{89.20}$   & $\textbf{90.85}$    & $\textbf{83.23}$ \\
    MC-Loss-V$2$                        & VGG$16$       & $65.20$             & $88.65$            & $90.53$             & $82.75$ \\
    MC-Loss minus $L_{div}$             & VGG$16$       & $64.52$             & $87.58$            & $89.55$             & $81.60$ \\
    MC-Loss minus $L_{dis}$             & VGG$16$       & $26.94$             & $79.75$            & $69.13$             & $38.53$ \\
    MC-Loss minus CWA                   & VGG$16$       & $63.36$             & $88.30$            & $89.34$             & $80.76$  \\
    \hline
    \end{tabular}%
  \label{tab:table_Ablation}%
\end{table*}%




\begin{figure*}[!t]
    \tiny
    \begin{center}

   \includegraphics[width=0.98\linewidth]{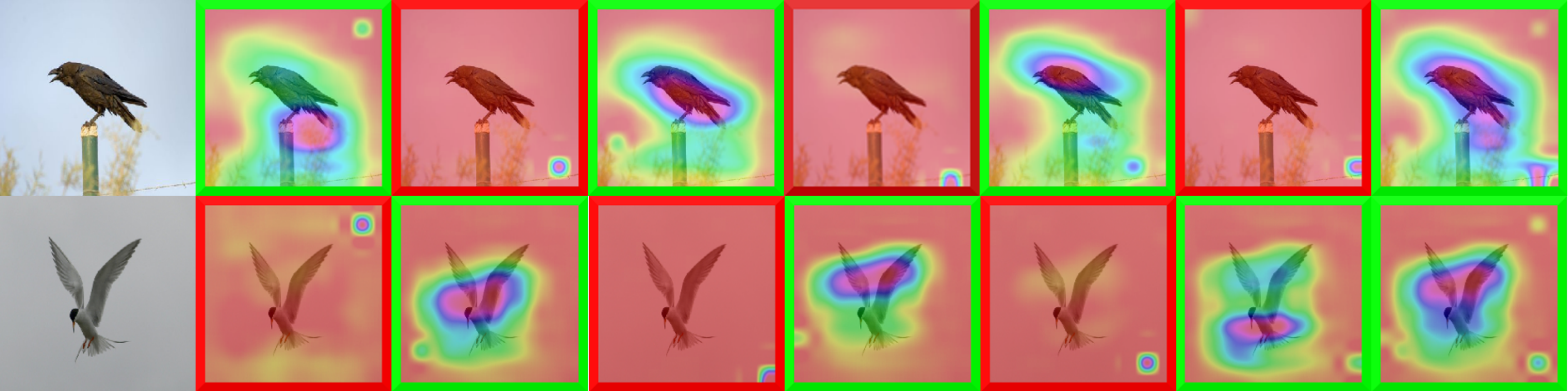}

    \end{center}
    \tiny
    \caption{Visualization of the localized regions returned from Grad-CAM~\cite{selvaraju2017grad} based on a VGG$16$ model (trained from scratch) optimized by the MC-Loss. The higher energy region denotes the more discriminative part in the image.}
    \label{fig:visualization_MC2}
\end{figure*}

\subsection{Comparisons with State-of-the-Art Methods}\label{SOTA}

Irrespective of the backbone networks, the proposed MC-Loss achieves a consistent improvement over the other state-of-the-art methods. Especially, the proposed MC-Loss achieves the best accuracy of $97.70\%$ on  Flowers-$102$ dataset. Detailed results are listed in Table~\ref{tab:SOTA_flowers102}. From Table~\ref{tab:SOTA}, it can be observed that our MC-Loss achieves best accuracies of $92.90\%$, $94.40\%$ on  FGVC-Aircraft and the Stanford Cars, respectively. Moreover, it obtains a competitive result on  CUB-$200$-$2011$ dataset. 

The component settings of the referred methods are also listed in Table~\ref{tab:SOTA}. In particular, A, B, C, D, and E denote stochastically initialized  classification layers (c-layers), pre-trained VGG16 with c-layers removed, pre-trained VGG19 with c-layers removed, pre-trained ResNet50 with c-layers removed, and pre-trained DenseNet161 with c-layers removed, respectively. While most of the methods modify their base architectures, the MC-Loss performs best on most datasets ~\emph{without any structural modification or adding extra parameters.} The only optimization procedure, Paired Confusion (PC), is based on the bilinear CNN (B-CNN) which is same as ours. The MC-Loss achieves remarkable improvement compared with PC on all four datasets. When the backbone of the MC-Loss is the pre-trained VGG$16$, the MC-Loss does not perform better on  CUB-$200$-$2011$ dataset,  one reason that  is due to the lack of feature channels. As mentioned in Table~\ref{tab:featuremaps}, with $512$ feature channels, $88$ classes on  CUB-$200$-$2011$ dataset have only two feature channels.  Since the birds on  CUB-$200$-$2011$ dataset have rich discriminative regions, it is difficult to obtain robust descriptions with insufficient number of feature channels. Hence, the performance is worse than some referred methods. For Stanford Cars dataset, although $112$ classes have two feature channels, the MC-Loss can still perform well due to the fact that the cars have less discriminative regions than the birds and the discriminative ability of the MC-Loss can compensate for the lack of feature channels.

\begin{figure*}[!t]
    \begin{center}
   \includegraphics[width=0.8\linewidth]{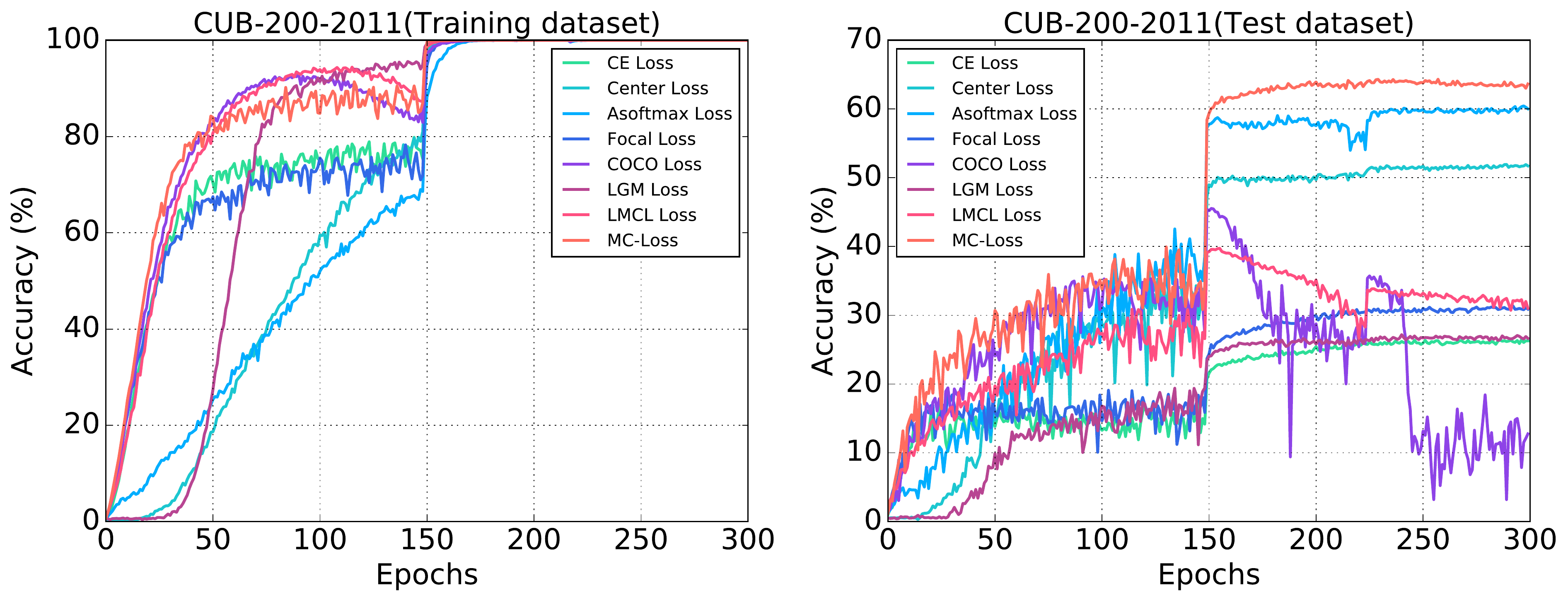}

    \end{center}
      \vspace{-5mm}
    \caption{The accuracies of the MC-loss and the other commonly used loss functions on the CUB-$200$-$2011$ dataset using the VGG$16$ as backbone.}
    \label{fig:acc_MC2}
      \vspace{-5mm}
\end{figure*}

\begin{figure*}[!t]
    \begin{center}
   \includegraphics[width=0.8\linewidth]{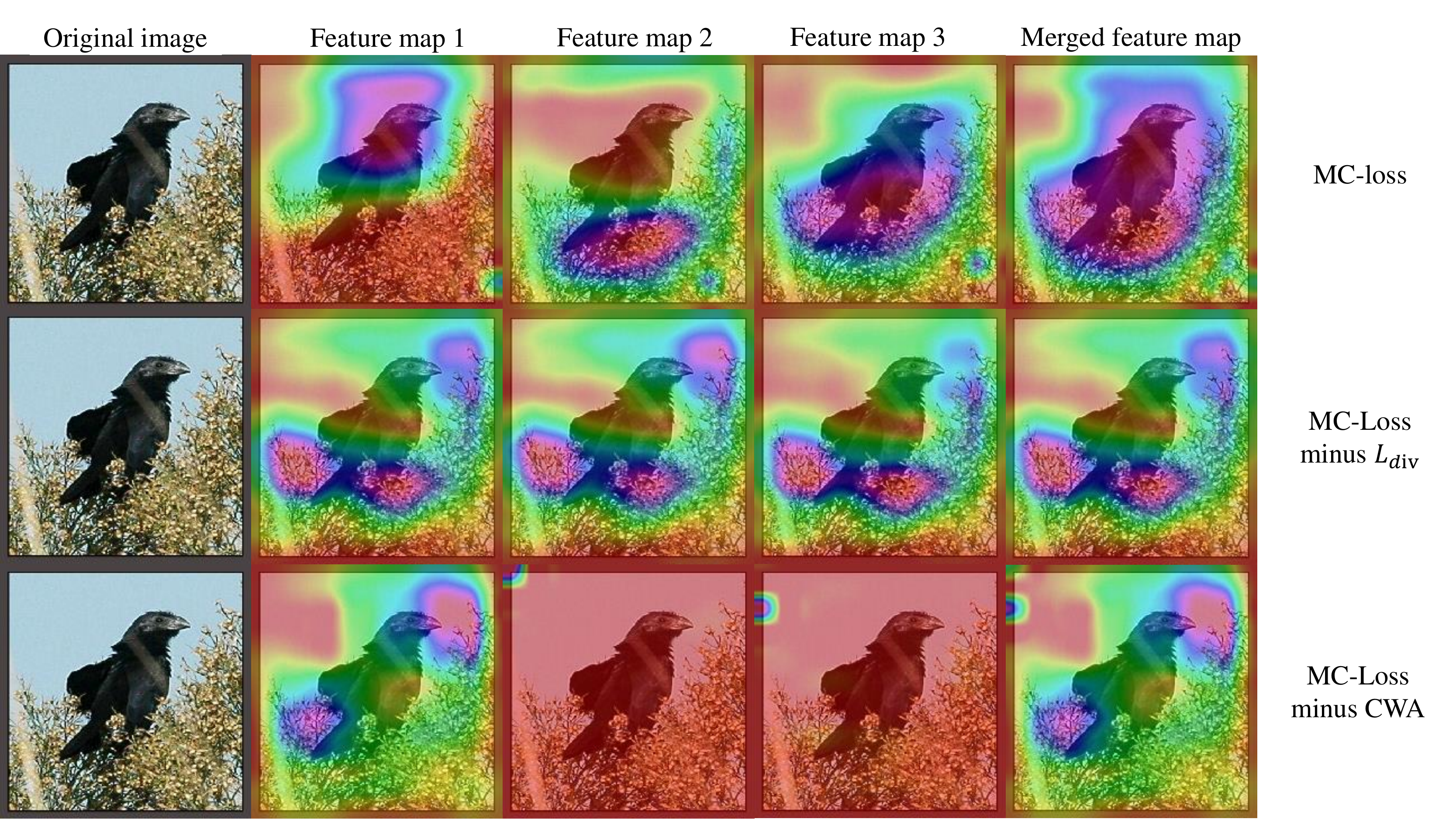}

    \end{center}
    \vspace{-5mm}
    \caption{Channel visualizations ($\xi=3$).  The first column represents the original image. The second to fourth columns show visualizations of the localization regions obtained from $3$ feature channels ($\xi=3$), respectively. The last column represents the visualizations of the merged localization regions of $3$ aforementioned feature channels.}
    \label{fig:visualization_MC}
      \vspace{-5mm}
\end{figure*}

\subsection{Ablation Study}\label{Ablation}

For ablation study, we train the backbone architecture (like the VGG$16$ or the ResNet$18$) from scratch using the loss function mentioned in Equation \ref{total_loss}, and we define number of output channels based on the requirement of assigning $\xi$ uniformly for every class, which is not possible with the pre-trained VGG$16$ because of its fixed channel outputs. We resize every image to $224\times224$. The learning rate of the complete network is initially set at $0.1$ and multiplied by $0.1$ at $150^{th}$ and $225^{th}$ epoch, while other settings are same with the earlier one. In addition, we set the hyper-parameters of the MC-Loss as $\mu$ =$1.5$ and $\lambda$=$10$. Although the  pre-trained VGG$16$ provides much better results, we have done this ablation study in order to justify the choice/potential of different hyper-parameters (like $\xi$) and different individual component of our loss function, where the backbone architecture is trained from scratch.

\noindent \textbf{Influence of $\xi$:}  In order to judge the influence of $\xi$ on the accuracy, we vary $\xi$ from $1$ to $6$ uniformly (for every class). Alongside, we also keep the $\xi$ assignment setup as detailed in Table \ref{tab:featuremaps} and term this as MC-Loss ($512$). From Table~\ref{tab:mis_match}, it can be interpreted that the MC-Loss ($\xi$=$1$) performs the worst and it signifies that only one discriminative region learned for each class is not enough for fine-grained image classification. The MC-Loss ($\xi$=$3$) achieves $3.71\%$ higher accuracy on CUB-$200$-$2011$ dataset compared to  MC-Loss ($512$) and thus it demonstrates that only two feature channels assigned to each class (recall from Table \ref{tab:featuremaps} that there are $88$ classes contain only two feature channels for each of them) is not sufficient to capture the discriminative information in bird's images. Increasing the $\xi$ value beyond $3$ decreases the performance along with the additional burden of the computational cost.
We speculate a drop in performance because when $\xi$ is large, the number of channels employed would exceed the number of useful ``parts", therefore, introducing redundant channels that are counter-effective. We also verified this through a few visualizations in Figure~\ref{fig:visualization_MC2}. The first column represents the original image. The second to seventh columns shows visualizations of the localization regions obtained from $6$ feature channels ($\xi=6$), respectively. The last column represents the visualizations of the merged localization regions of $6$ aforementioned feature channels. Red boxes: redundant channels; green boxes: channels that exhibit localized regions. Overall, it is to be noted that the number of feature channels has remarkable influence on the classification performance and accuracy is optimum when $\xi$ equals to $3$. Therefore, had we been able to train a new VGG$16$ model on ImageNet dataset with sufficient number of output channels, such that $\xi$ can be set to $3$ uniformly for every class in the fine-grained dataset, it is expected that the MC-Loss optimized on the top that pre-trained model could have performed better than other methods for fine-grained classification.

\noindent \textbf{Comparison with other loss-functions:} Table~\ref{tab:table_loss_function} shows the comparison between the proposed MC-Loss and other commonly used loss functions, on the  four widely used fine-grained image classification datasets. Results on the left and right hands of the slashes in the table are for the VGG$16$ and  the ResNet$18$ respectively. Using the VGG$16$ model as the feature extractor, the proposed MC-Loss achieves the best accuracies of $65.98\%$, $89.20\%$, $90.85\%$, and $83.23\%$ on CUB-$200$-$2011$, FGVC-Aircraft, Stanford Cars, and Flowers-$102$ datasets, respectively. While using ResNet$18$ model as the feature extractor, the proposed MC-Loss still obtains the best performance on four fine-grained image classification datasets. In summary, the proposed MC-Loss outperforms all the compared methods on all the four fine-grained image classification datasets for both the VGG$16$ and the ResNet$18$ base networks. 
{Meanwhile, Figure~\ref{fig:acc_MC2} demonstrates the accuracy curves of the MC-Loss and the  other commonly used loss functions on the CUB-$200$-$2011$ dataset. 
From  Figure~\ref{fig:acc_MC2}, the MC-Loss exhibits improved optimisation characteristics and produces consistent gains in performance which are sustained throughout the training process.}


\noindent \textbf{Visualization:} In order to illustrate the advantages of the proposed MC-Loss intuitively, we applied the Grad-CAM~\cite{selvaraju2017grad} to implement the visualization for the feature channels. The first row of Figure~\ref{fig:visualization_MC} shows the most discriminative regions proposed by the  VGG$16$ model while trained using the complete MC-Loss. It can be observed that the three feature channels that corresponds to a specified bird class focus on different discriminative regions, \emph{e.g.} head, feet, wings, and body. Meanwhile, the second row of Figure~\ref{fig:visualization_MC} shows the most discriminative regions while using only discriminality component alone in the MC-loss. We can observe  that if we do not use the diversity component in the MC-Loss, the three feature channels learned by the VGG$16$  model tend to be similar to each other. This indicates that the learned feature channels cannot focus on different discriminative regions in absence of the diversity component, which reduces its ability in fine-grained image classifications. The last row of Figure~\ref{fig:visualization_MC} shows an example of the most discriminative regions predicted by the VGG$16$ model optimized by the MC-loss without channel-wise attention operation. It can be clearly interpreted that if the channel-wise attention module is removed, only one of the three feature channels represents the correct discriminative region. The other two feature channels, although are different from each other, do not necessarily learn any discriminative information. 

The quantitative comparisons about the aforementioned phenomenon are listed in Table~\ref{tab:table_Ablation}. We can observe that, if we use only the discriminality component of the MC-Loss, the classification accuracies drop by $1.46\%$, $1.62\%$, $1.30\%$, and $1.63\%$ on CUB-$200$-$2011$, FGVC-Aircraft, Stanford Cars, and Flowers-$102$ datasets, respectively. Furthermore, if we remove the channel-wise attention in the discriminality component in the MC-Loss, the accuracies will be decreased by $2.60\%$, $0.90\%$, $1.51\%$, and $2.47\%$, respectively. Alternatively, in contrast to Equation~\ref{equ:ldiv} one could only consider the channel group that belongs to the ground-truth class. In particular, Equation~\ref{equ:ldiv} could be replaced as ${L}_{div\_v2}(\mathbf{F})=h(\mathbf{F}_i)$. In Table~\ref{tab:table_Ablation}, we report the performance of this design as MC-Loss-V$2$, while the rest of the things remain unchanged. We can see that the classification accuracy drops significantly. The reason is that while using our proposed diversity loss component all the channel groups influence each other during  training, but this ${L}_{div\_v2}$  only considers the diversity of a channel group belonging to the ground truth class during training. In other words, if we only consider one channel group during training, the other groups of channels might lose the diversity. Intuitively, Equation~\ref{equ:ldiv} is being able to cultivate cross-group/class information, which essentially helps the final classification. These results are consistent with the analysis about the visualizations in Figure~\ref{fig:visualization_MC}.

\subsection{MC-Loss with Soft Channel Label}\label{WMC}

\begin{figure}[!t]
  \begin{center}
    \includegraphics[width=1.0\linewidth]{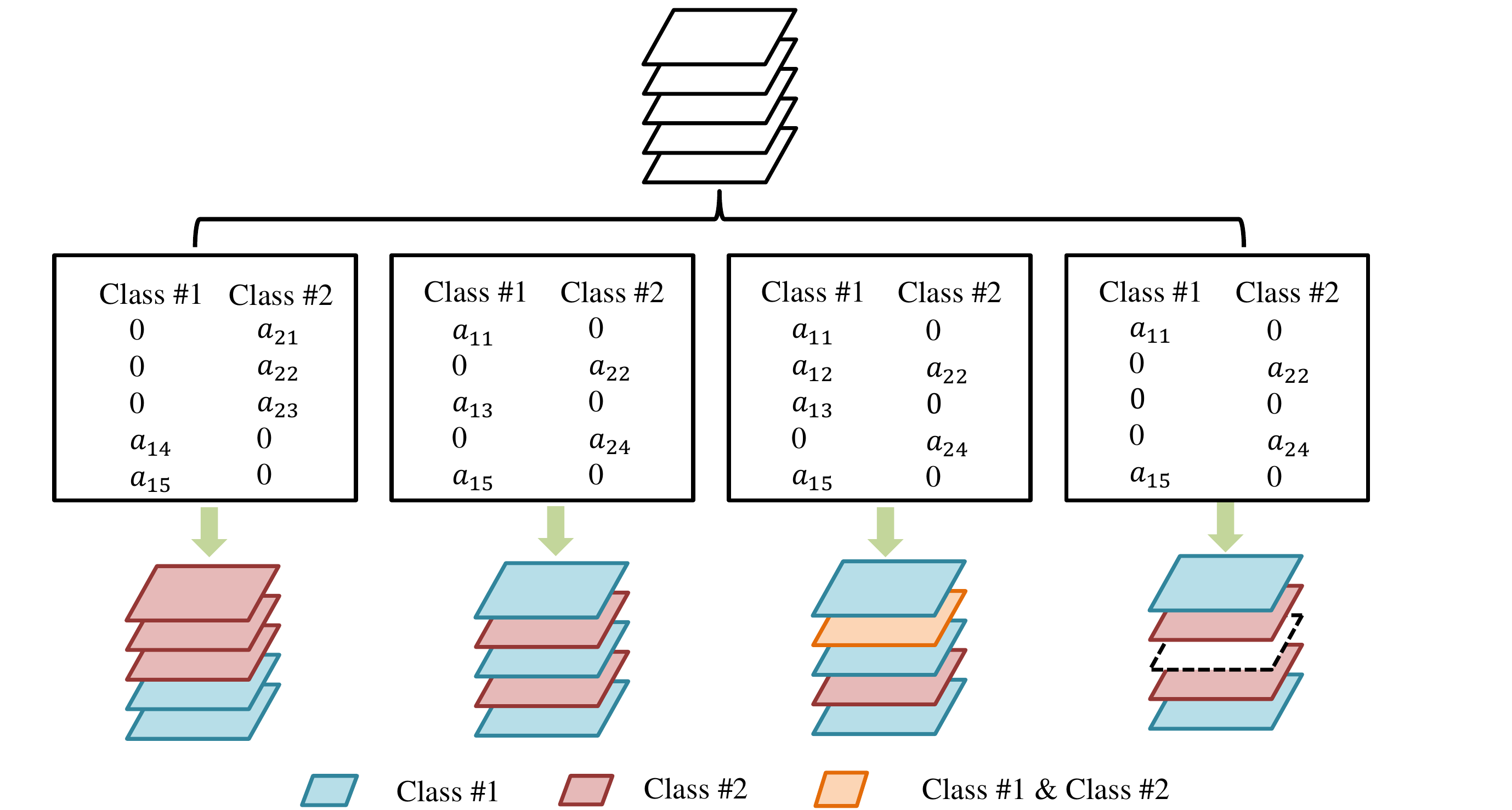}
  \end{center}
  \vspace{-5mm}
  \caption{Explanation of the soft channel label used in our MC-Loss with two categories as an example. The solid line boxes show four learned soft channel labels, and $a_{pq}$ represents a number greater than $0$ and less than $1$, where $p$ indicates the category,  $q$ indicates the the total number of classes. 
  In the first box, the learned soft channel labels make the channels contained in the feature group of each category aggregate with their neighborhoods, which obeys the original settings. 
  In the second box, the corresponding feature group of each category are scattered in the feature channels.
  In the third box, the second channel is shared by both categories.
  In the last box, the third channel is not assigned to any specific categories.
  ~\textbf{Please refer to Section~\ref{channelsoftlabel} for details.}
  }
  \label{fig:soft}
    \vspace{-5mm}
\end{figure}

\begin{table*}[!t]
  \centering
  \small
  \caption{Influence of the soft channel labels on three fine-grained image classification datasets using the pre-trained ResNet$50$.}
    \begin{tabular}{|c|c|c|c|c|c|}
    \hline
    Method                                 & Base Model   & CUB-$200$-$2011$     & FGVC-Aircraft     & Stanford Cars \\
    \hline
    \hline
    MC-Loss&                               ResNet$50$     & $87.3$               & $92.6$            & $93.7$  \\          
    MC-Loss with soft channel label&       ResNet$50$     & $87.8$               & $92.9$            & $94.1$  \\       
    \hline
    \end{tabular}%
  \label{tab:soft}%
\end{table*}%

\begin{table*}[!t]
  \centering
  \caption{Examples of the soft channel labels learned by the  ResNet$50$ (the channel number is fixed at $2048$), showing the $10$ most important channels for a particular category, the importance of which gradually decreases from left to right. Examples of the first five categories of the  CUB-$200$-$2011$ dataset are provided.}
    \begin{tabular}{|c|cccccccccc|}
    \hline
          & \multicolumn{10}{c|}{Top 10 important channels}\\
\cline{2-11}    Category   & $1$     & $2$     & $3$     & $4$     & $5$     & $6$     & $7$     & $8$     &$9$      & $10$ \\
    \hline
    \hline
    1                   & $175$   & $1119$  & $1844$  & $628$   & $833$   & $815$   & $819$   & $1938$  & $1837$  & $1135$ \\
    2                   & $175$   & $1119$  & $1577$  & $628$   & $1676$  & $1837$  & $1844$  & $819$   & $1135$  & $1838$ \\
    3                   & $775$   & $1188$  & $1914$  & $668$   & $465$   & $1765$  & $1066$  & $591$   & $1053$  & $670$ \\
    4                   & $257$   & $1776$  & $760$   & $235$   & $1739$  & $753$   & $1212$  & $269$   & $1215$  & $1740$ \\
    5                   & $1933$  & $257$   & $746$   & $1066$  & $1217$  & $1219$  & $1063$  & $273$   & $1740$  & $1739$ \\
    \hline
    \end{tabular}%
  \label{tab:channels}%
\end{table*}%

In order to make the proposed loss function more flexible and adaptive to the structures of the networks, especially to adapt the number of channels for network-extracted features, we propose the soft channel label in this section, which are used to assign channels to specific categories. Unlike the original settings, channels belonging to a particular category are no longer required to be aggregated into groups with their neighborhoods. In addition, the number of channels assigned to each category is uneven  in this case, which facilitates the learning of more discriminative and diverse patterns  in  the difficult categories. 

The learnable soft channel labels determine the channel groups for each category. In order to enable the ability of the model to learn soft channel labels, we use the SE-block ~\cite{hu2018squeeze} which can learn the channel attention weight for each channel. During the training process, the soft channel labels are constrained so that those belonging to the same category have high similarity, while those belonging to different classes have low similarity. 
It should be emphasized that the optimization objective of the model in (\ref{total_loss}) is unchanged, except that the grouping of features has been modified. In the following content, we will introduce how the soft channel labels are learned in detail and the penalties are imposed on them.

\subsubsection{Soft Channel Label}\label{channelsoftlabel}

As is well known, the SE-block can learn the channel attention weight for each channel. Therefore, we use the SE-block to learn the soft channel  labels for each category. The weights belonging to one sample can be defined by





\begin{small}
\begin{equation}
w_{i,j}=\left[t_{1}^{(i,j)},\ t_{2}^{(i,j)},...,t_{k-1}^{(i,j)},t_{k}^{(i,j)}\right]^\text{T},
    \label{mcloss}
\end{equation}
\end{small}


\noindent where $t_{k}^{i,j} \in (0, 1)$ indicates the importance of the specific channel, $k = 1,2,...,N$ ($N$ is the total number of channels) indicates the channel index, $i$ indicates the category, $i =1,2,...,C $ ($C$ is the total number of classes), and $j$ indicates the sample index belonging to the $i^{th}$  category, $j = 1,2,\cdots,J_i$  ($J_i$ is the total number of samples belonging to the $i^{th}$  category).

In addition, during the training process, each batch  contains different categories of data, and the sample number in the batch of each category is inconsistent. The soft channel labels of the samples belonging to the same category are expressed as

\begin{small}
\begin{equation}
W_{i} = \left[w_{i,1}, w_{i,2},...,w_{i,J_i-1},w_{i,J_i}\right]^\text{T},
    \label{wi}
\end{equation}
\end{small}

\noindent where the dimension of the $W_{i}$ is $ J_i\times N$.

It can be found in Section~\ref{Methods} that the CCMP is an effective way to measure the difference between feature channels. However, the CCMP can  be only  used on channels belonging to one mere sample, rather than different samples, which is not suitable in this case. Therefore, we use a similar way to measure the similarity between the soft channel labels belonging to the same class, and propose a loss component $L_{intra}$ defined as


\begin{small}
\begin{equation}
L_{intra}=\frac{1}{C} \sum_{i=1}^{C}\sum_{k=1}^{N}\max_{j=1,\cdots,J_i}\left(W_i\right).
    \label{intra}
\end{equation}
\end{small}



Intuitively, the soft channel labels belonging to different classes should be different when we apply them to assign the feature channels. Therefore, we also use the CCMP to measure the similarity between soft channel labels belonging to the different classes, and propose a loss component $L_{inter}$ defined as

\begin{small}
\begin{equation}
L_{inter}=- \sum_{k=1}^{N}\max_{i=1,\cdots,C} \max_{j=1,\cdots,J_i}\left(W_i\right) .
    \label{intra}
\end{equation}
\end{small}

Thus, the total loss function of the whole network can be defined as:

\begin{small}
\begin{equation}
  Loss(\mathbf{F})=L_{CE}(\mathbf{F})+\mu\times L_{MC}(\mathbf{F})+L_{intra} + L_{inter}.
 \label{total_loss}
\end{equation}
\end{small}

Figure~\ref{fig:soft} shows an illustration to illustrate the key idea of how the learned soft channel labels assign the feature channels to all classes and obtain a set of feature channels that are class-aligned.

\subsubsection{Experiments}\label{softExperiment}
We evaluate the proposed MC-Loss on the three aforementioned fine-grained  image classification data. Except that channel assignment is no longer set manually but can be learned by the model itself, other settings remain the same. Furthermore, we set the hyper-parameters of the MC-Loss as $\mu$ =$0.5$ and $\lambda$=$10$.

Table~\ref{tab:soft} lists the comparison between the proposed MC-Loss and the MC-Loss with the soft channel labels. Using the pre-trained ResNet$50$ model as the feature extractor, the proposed MC-Loss with the soft channel label achieves the better accuracies of $87.8\%$, $92.9\%$, and $94.1\%$ on CUB-$200$-$2011$, FGVC-Aircraft, and Stanford Cars datasets, respectively. In summary, the proposed soft channel labels make the MC-Loss more  flexible and more adaptive to the structure of the network, especially the number of channels for network-extracted features, and have no change on the distribution of features, more easy to fine-tune.

In order to illustrate the advantages of the soft channel label, we show  the channels assigned  for each category based on the learned soft channel labels in Table~\ref{tab:channels}. It can be observed that the soft channel label-supervised channels assigned  to different categories according to the soft channel labels are significantly different, and some channels are shared. 
For example, for the $5^{th}$ category, the most important channel is the $1933^{th}$ channel and the second most important one is the $1257^{th}$ channel.
\section{Conclusions}\label{Conclusions}

In this paper, we show that it is possible to learn discriminate localized part features for fine-grained classification, with a single loss. The proposed mutual-channel loss (MC-Loss) can effectively drive the feature channels to be more discriminative and focusing on various regions, without the need of fine-grained bounding-box/part annotations. We show that our loss can applied to different network architectures, and does not introduce any extra parameters in doing so. Experiments on all four fine-grained classification datasets confirm the superiority of the MC-Loss. In the future, we will investigate means of automatically searching for $\xi$, without necessarily introducing considerably more network parameters. We will also look into applying the MC-Loss to other tasks that rely on local and discriminative regions, and extending it to work across different modalities (~\emph{e.g.}, for fine-grained sketch-based image retrieval).


%





\ifCLASSOPTIONcaptionsoff
  \newpage
\fi



%

{\small
\bibliographystyle{ieee}
\bibliography{main}
}

\end{document}